\documentclass[10pt,twocolumn,letterpaper]{article}

\usepackage{cvpr}              

\def\note{$^{\dag}$}
\def\nnote{$^{\ddag}$}

\definecolor{background_gray}{gray}{0.84}

\definecolor{cell1_r0_c0}{rgb}{0.583,0.300,0.385}
\definecolor{cell1_r0_c1}{rgb}{0.897,0.586,0.528}
\definecolor{cell1_r0_c2}{rgb}{0.986,0.849,0.779}
\definecolor{cell1_r0_c3}{rgb}{0.943,0.962,0.972}
\definecolor{cell1_r0_c4}{rgb}{0.477,0.694,0.830}
\definecolor{cell1_r0_c5}{rgb}{0.772,0.877,0.930}
\definecolor{cell1_r1_c0}{rgb}{0.980,0.967,0.960}
\definecolor{cell1_r1_c1}{rgb}{0.935,0.958,0.970}
\definecolor{cell1_r1_c2}{rgb}{0.826,0.905,0.945}
\definecolor{cell1_r1_c3}{rgb}{0.826,0.905,0.945}
\definecolor{cell1_r1_c4}{rgb}{0.980,0.971,0.965}
\definecolor{cell1_r1_c5}{rgb}{0.985,0.946,0.924}
\definecolor{cell1_r2_c0}{rgb}{0.989,0.928,0.893}
\definecolor{cell1_r2_c1}{rgb}{0.628,0.795,0.885}
\definecolor{cell1_r2_c2}{rgb}{0.344,0.490,0.647}
\definecolor{cell1_r2_c3}{rgb}{0.317,0.438,0.574}
\definecolor{cell1_r2_c4}{rgb}{0.344,0.490,0.647}
\definecolor{cell1_r2_c5}{rgb}{0.799,0.891,0.938}
\definecolor{cell1_r3_c0}{rgb}{0.923,0.952,0.968}
\definecolor{cell1_r3_c1}{rgb}{0.980,0.971,0.965}
\definecolor{cell1_r3_c2}{rgb}{0.982,0.958,0.944}
\definecolor{cell1_r3_c3}{rgb}{0.894,0.938,0.963}
\definecolor{cell1_r3_c4}{rgb}{0.628,0.795,0.885}
\definecolor{cell1_r3_c5}{rgb}{0.711,0.846,0.912}
\definecolor{cell1_r4_c0}{rgb}{0.968,0.973,0.976}
\definecolor{cell1_r4_c1}{rgb}{0.853,0.918,0.953}
\definecolor{cell1_r4_c2}{rgb}{0.620,0.790,0.882}
\definecolor{cell1_r4_c3}{rgb}{0.637,0.800,0.888}
\definecolor{cell1_r4_c4}{rgb}{0.704,0.843,0.910}
\definecolor{cell1_r4_c5}{rgb}{0.758,0.870,0.926}
\definecolor{cell1_r5_c0}{rgb}{0.964,0.971,0.975}
\definecolor{cell1_r5_c1}{rgb}{0.902,0.942,0.964}
\definecolor{cell1_r5_c2}{rgb}{0.711,0.846,0.912}
\definecolor{cell1_r5_c3}{rgb}{0.704,0.843,0.910}
\definecolor{cell1_r5_c4}{rgb}{0.826,0.905,0.945}
\definecolor{cell1_r5_c5}{rgb}{0.991,0.878,0.817}

\definecolor{cell2_r0_c0}{rgb}{0.995,0.901,0.846}
\definecolor{cell2_r0_c1}{rgb}{0.911,0.946,0.966}
\definecolor{cell2_r0_c2}{rgb}{0.779,0.880,0.932}
\definecolor{cell2_r0_c3}{rgb}{0.418,0.616,0.791}
\definecolor{cell2_r0_c4}{rgb}{0.403,0.597,0.781}
\definecolor{cell2_r0_c5}{rgb}{0.383,0.565,0.752}
\definecolor{cell2_r1_c0}{rgb}{0.981,0.820,0.742}
\definecolor{cell2_r1_c1}{rgb}{0.927,0.954,0.969}
\definecolor{cell2_r1_c2}{rgb}{0.874,0.929,0.959}
\definecolor{cell2_r1_c3}{rgb}{0.806,0.894,0.939}
\definecolor{cell2_r1_c4}{rgb}{0.731,0.856,0.918}
\definecolor{cell2_r1_c5}{rgb}{0.480,0.699,0.833}
\definecolor{cell2_r2_c0}{rgb}{0.841,0.471,0.467}
\definecolor{cell2_r2_c1}{rgb}{0.894,0.938,0.963}
\definecolor{cell2_r2_c2}{rgb}{0.451,0.660,0.813}
\definecolor{cell2_r2_c3}{rgb}{0.671,0.822,0.899}
\definecolor{cell2_r2_c4}{rgb}{0.458,0.670,0.818}
\definecolor{cell2_r2_c5}{rgb}{0.368,0.536,0.712}
\definecolor{cell2_r3_c0}{rgb}{0.984,0.837,0.765}
\definecolor{cell2_r3_c1}{rgb}{0.806,0.894,0.939}
\definecolor{cell2_r3_c2}{rgb}{0.611,0.784,0.879}
\definecolor{cell2_r3_c3}{rgb}{0.350,0.502,0.663}
\definecolor{cell2_r3_c4}{rgb}{0.323,0.449,0.590}
\definecolor{cell2_r3_c5}{rgb}{0.314,0.432,0.566}
\definecolor{cell2_r4_c0}{rgb}{0.980,0.814,0.735}
\definecolor{cell2_r4_c1}{rgb}{0.985,0.943,0.919}
\definecolor{cell2_r4_c2}{rgb}{0.984,0.952,0.934}
\definecolor{cell2_r4_c3}{rgb}{0.586,0.768,0.870}
\definecolor{cell2_r4_c4}{rgb}{0.569,0.757,0.864}
\definecolor{cell2_r4_c5}{rgb}{0.329,0.461,0.607}
\definecolor{cell2_r5_c0}{rgb}{0.946,0.697,0.614}
\definecolor{cell2_r5_c1}{rgb}{0.583,0.300,0.385}
\definecolor{cell2_r5_c2}{rgb}{0.986,0.940,0.913}
\definecolor{cell2_r5_c3}{rgb}{0.725,0.853,0.916}
\definecolor{cell2_r5_c4}{rgb}{0.484,0.704,0.835}
\definecolor{cell2_r5_c5}{rgb}{0.362,0.525,0.695}

\definecolor{cvprblue}{rgb}{0.21,0.49,0.74}
\usepackage[pagebackref,breaklinks,colorlinks,allcolors=cvprblue]{hyperref}

\usepackage{makecell}
\usepackage{multirow}
\usepackage{hyperref}
\usepackage{xcolor}
\usepackage{colortbl}
\usepackage{diagbox}
\def\paperID{ } 
\def\confName{CVPR}
\def\confYear{2026}

\title{IncreFA: Breaking the Static Wall of Generative Model Attribution}

\author{Haotian Qin$^1$, Dongliang Chang$^1$\thanks{Corresponding author.} , Yueying Gao$^1$, Yuexuan Tan$^1$, Lei Chen$^2$ and Zhanyu Ma$^1$\\
$^1$ {\small School of Artificial Intelligence, Beijing University of Posts and Telecommunications, China} \\
$^2$ {\small Tsinghua University, China} \\
{\tt\small\{qinhaotian, changdongliang, gaoyueying, tanyuexuan, mazhanyu\}@bupt.edu.cn} \\
{\tt\small leichenthu@tsinghua.edu.cn} \\
}

\begin{document}
\maketitle

\begin{abstract}
As AI generative models evolve at unprecedented speed, image attribution has become a moving target. New diffusion, adversarial and autoregressive generators appear almost monthly, making existing watermark, classifier and inversion methods obsolete upon release. The core problem lies not in model recognition, but in the inability to adapt attribution itself.
We introduce \textbf{IncreFA}, a framework that redefines attribution as a \textbf{structured incremental learning} problem, allowing the system to learn continuously as new generative models emerge. IncreFA departs from conventional incremental learning by exploiting the hierarchical relationships among generative architectures and coupling them with continual adaptation. It integrates two mutually reinforcing mechanisms: (1) \textbf{Hierarchical Constraints}, which encode architectural hierarchies through learnable orthogonal priors to disentangle family-level invariants from model-specific idiosyncrasies; and (2) a \textbf{Latent Memory Bank}, which replays compact latent exemplars and mixes them to generate pseudo-unseen samples, stabilising representation drift and enhancing open-set awareness.
On the newly constructed \textbf{Incremental Attribution Benchmark (IABench)} covering 28 generative models released between 2022 and 2025, IncreFA achieves state-of-the-art attribution accuracy and 98.93\% unseen detection under a temporally ordered open-set protocol. Code will be available at \url{https://github.com/Ant0ny44/IncreFA}.
\end{abstract}

\vspace{-0.1in}
\section{Introduction}
\label{sec:intro}

\begin{figure}[ht]
    \centerline{\includegraphics[width=0.95\columnwidth]{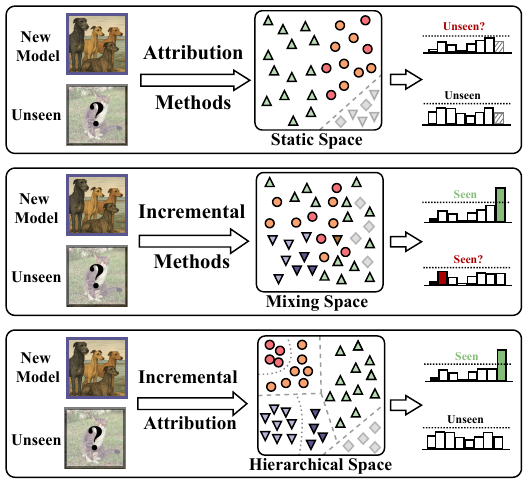}}
    \vspace{-0.15in}
    \caption{\textbf{Hierarchical Incremental Attribution.} When new generative models emerge, IncreFA rapidly adapts without forgetting previous ones by learning hierarchically orthogonal priors and replaying latent memories.}
    \vspace{-0.25in}
    \label{fig:intro}
\end{figure}

The rapid proliferation of AI generative models has reshaped the visual content landscape \cite{gao2023self, fakereasoning2025}. While these models enable creative innovation \cite{demofusion2024}, they also introduce serious risks such as copyright infringement, content forgery and unauthorised image use \cite{where2023}. Reliable attribution, which identifies the originating model of a generated image, is essential for ensuring accountability and maintaining public trust in digital media.

Recent progress in image attribution has explored diverse directions, ranging from watermark embedding to feature-based attribution analysis and representation-driven tracing. Early approaches relied on imperceptible watermarks to tag model outputs \cite{watermark1996,watermark2000,reversible2009,stamp2020}, but these require cooperation from model developers and often compromise visual fidelity. Later methods shifted towards classifier-based recognition that learns to distinguish the visual fingerprints of different generators \cite{reversible2009,stamp2020,attribution2019,openattribution2021}, yet these models quickly become outdated as new architectures emerge. More recent studies leverage latent inversion \cite{where2023,trace2024} or large-scale pretrained encoders \cite{defake2023} such as CLIP \cite{clip2021} and DINO \cite{dino2023} to match high-level representations between generated and real images. While some classifier-based approaches \cite{openattribution2021,dna2022} have begun to incorporate open-set handling through confidence calibration or thresholding, most attribution pipelines still rely on a closed-set assumption tied to specific known generators. In practice, new diffusion, adversarial and autoregressive models appear almost monthly, and even subtle architectural variants within a family can produce highly confusable traces, creating a moving target beyond static attribution.

Rather than a one-time classification task, attribution should evolve with the generative landscape. The challenge lies not in recognising known generators, but in continuously attributing newly released ones without erasing past knowledge \cite{coutinualsurvey2024}. We therefore reformulate the task as incremental attribution, where the model continuously learns to attribute newly released generators while retaining the ability to recognise previous ones. This setting reflects real-world model evolution and raises two central challenges: retaining family-level invariants while separating model-specific idiosyncrasies, and distinguishing seen generators from an expanding open space with minimal false alarms.

Conventional class-incremental learning methods have been widely explored in image classification \cite{compreincresurvey2024}, but they rely on assumptions that do not hold here. Typical benchmarks assume balanced class splits and separable domains \cite{simplecil2025,l2p2022}, whereas generative models form overlapping distributions and share inherited patterns across versions. When a new diffusion variant inherits most of its parent’s statistics, gradient updates drift toward previously occupied regions, leading to negative transfer even with replay. Moreover, existing methods focus purely on task retention and rarely consider open-set recognition \cite{l2p2022,mos2025,tuna2025}, which is critical for detecting unseen generators. These differences call for an attribution-specific formulation that integrates hierarchical structure awareness with continual adaptation.

We introduce \textbf{IncreFA}, a framework that breaks the static assumption of previous attribution and redefines it as a structured incremental learning problem. Unlike conventional incremental methods that treat tasks as independent, IncreFA exploits the inherent hierarchy among generative models and learns from their interrelated latent structures. 
Generative models are not isolated classes but hierarchical lineages, and this insight motivates a new constraint design that respects their architectural ancestry. 
It builds upon two complementary mechanisms that jointly enable continual, structured and open-set attribution.  
(1) \textbf{Hierarchical Constraints} explicitly encode the architectural hierarchy of generative models by learning orthogonal priors across coarse and fine levels. These priors are instantiated as learnable targets that anchor projected latents, enforcing angular separation across families while promoting knowledge sharing within related models.  
(2) \textbf{Latent Memory Bank} preserves compact latent exemplars instead of raw images and performs random feature mixing to generate pseudo-unseen samples. Latent mixing drives samples toward low-density regions near decision boundaries, approximating the open-set boundary and providing hard negatives for calibration. These two modules stabilise representation drift and enable attribution knowledge to evolve without sacrificing discrimination or adaptability.

This design yields a compact, data-efficient and continually extensible attribution framework. On the newly constructed \textbf{Incremental Attribution Benchmark (IABench)} covering 28 generative models released between 2022 and 2025, under a temporally ordered open-set protocol where the latest families are withheld as unknowns, IncreFA achieves state-of-the-art performance and consistently exceeding 98\% unseen detection across all settings, confirming the effectiveness of our formulation for continual and open-set attribution.

\section{Related Work}
\label{sec:related_work}

\subsection{AI-Generated Image Attribution}
AI-generated image analysis has traditionally focused on detection, which aims to distinguish real from synthetic content through frequency-domain analysis \cite{freqnet2024,fatformer2024,FF++2019}, spatial artefact modelling \cite{cnn2020,l2p2022,cospy2025,chameleon2025,unifd2023,exposingfake2024}, or multimodal representation learning \cite{defake2023,qin2025}. While such methods have achieved strong generalisation in identifying authenticity, attribution presents a more demanding challenge because it requires determining which generative model produced an image rather than simply deciding whether it was generated.

Existing attribution approaches can be broadly divided into three categories: watermark-based, classifier-based, and inversion-based. Watermarking methods embed imperceptible identifiers into generated images \cite{watermark1996,watermark2000,reversible2009,artificial2021}, but they depend on the cooperation of model developers and may compromise visual fidelity. Inversion-based methods \cite{where2023,attibutiondeepfakes2020,source2019} reconstruct latent codes to match a suspect model, but this process usually requires white-box access to model parameters, which is rarely feasible in practice. Classifier-based methods have therefore become the dominant paradigm, training discriminative models to learn visual fingerprints across different generators \cite{defake2023,dna2022,pose2023,repmix2022}. Some recent studies include open-set handling through confidence calibration or distance thresholding \cite{pose2023}, yet most pipelines still rely on a closed taxonomy that assumes known generators. In real-world scenarios, diffusion, adversarial and autoregressive models evolve rapidly, producing overlapping and inherited feature distributions that make static attribution increasingly impractical. This situation motivates a new perspective where attribution itself must evolve together with generative models. Inspired by several fine-grained studies \cite{fgpro2020,fgornot2021}, we aim to conduct hierarchical and fine-grained attribution method.

\subsection{{Incremental Learning for Attribution}}
Class-incremental learning (CIL) provides a potential direction for achieving continual adaptation, allowing models to learn new categories while retaining knowledge of previously learned ones \cite{compreincresurvey2024}. Existing CIL methods can be grouped into four categories: replay-based \cite{rainbow2021,cored2021,facecontinual2025}, regularisation-based \cite{overcoming2018}, optimiser-based \cite{layerwise2021,adaptive2022,lrorth2020}, and representation-based methods \cite{icarl2017,mimicking2022,class2022,emp2023}. These methods have achieved strong performance in object recognition tasks. However, their underlying assumptions do not hold for model attribution. In attribution, the classes are not independent because generative models form hierarchical and overlapping lineages, and new versions often inherit the visual statistics of their predecessors. This inheritance leads to feature drift and negative transfer even when replay mechanisms are used. In addition, most incremental learning frameworks ignore open-set recognition, which is crucial for detecting images from unseen generators.

To address these issues, we formulate \textbf{incremental attribution}, which integrates continual learning with the hierarchical structure of generative models. This formulation captures both family-level invariants and model-specific variations, enabling attribution to evolve in step with the generative landscape. The following section presents its formal definition and design rationale.

\section{Problem Formulation}

\subsection{Task Definition}
The goal of image attribution is to identify the originating model of a generated image. Given an input image \(x\) and a set of candidate generative models \(\mathcal{M} = \{\mathcal{M}_1, \mathcal{M}_2, \dots, \mathcal{M}_n\}\), the attribution model \(F\) predicts the most likely source as
\begin{equation}
\hat{y}=\arg\max_i F\big(x\big),\quad \mathcal{M}_i\in\mathcal{M}.
\label{Eq:goal}
\end{equation}
In practice, the image may either be real or produced by an unseen model that has not appeared during training. The decision space is therefore extended to \(\{R, \mathcal{M}_1, \dots, \mathcal{M}_n, \mathcal{M}_{\text{unseen}}\}\). This open and evolving setting departs from traditional closed-set recognition and motivates a formulation that can adapt continually to new generators.

\subsection{Hierarchical Structure of Generative Models}
Generative models are not independent classes but belong to hierarchical families that share architectural ancestry. For instance, diffusion models of different versions often inherit latent statistics and stylistic priors from their predecessors. This lineage produces overlapping feature distributions that violate the separability assumption in conventional incremental learning.

We define the attribution feature space as a structured hierarchy composed of \(K\) coarse-level families, each containing \(N_k\) fine-grained models. Let \(\mathcal{Z}\) denote the latent representation space extracted from a pretrained backbone, and \(\mathcal{P}_k \subset \mathbb{R}^d\) represent the subspace assigned to the \(k\)-th family. The hierarchy is defined as
\begin{equation}
\mathcal{Z}' = \bigoplus_{k=1}^{K} \mathcal{P}_k, \quad \mathcal{P}_i \perp \mathcal{P}_j \text{ for } i \neq j,
\label{eq:hierarchy}
\end{equation}
where \(\bigoplus\) denotes the direct-sum decomposition of the latent space and $\mathcal{Z} =  \mathcal{Z}'\bigoplus\mathcal{Z}_{real}$. Each \(\mathcal{P}_k\) groups multiple related models \(\{\mathcal{M}_{k,1}, \dots, \mathcal{M}_{k,N_k}\}\) that share family-level invariants. This formulation preserves inter-family orthogonality while enabling intra-family knowledge sharing, providing the structural foundation for our hierarchical constraint design.
\vspace{-0.1in}
\begin{figure}[ht]
    \centerline{\includegraphics[width=1.0\columnwidth]{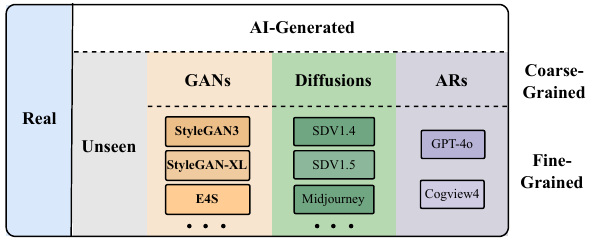}}
    \vspace{-0.05in}
    \caption{\textbf{Hierarchical Relation in Attribution.} Images are first categorised as real, fake, or unseen. Fake images are further divided into GAN, diffusion, and autoregressive families, each occupying an orthogonal subspace. Models within the same family remain correlated to preserve shared visual statistics.}
    \vspace{-0.15in}
    \label{fig:hierarchical_formulation}
\end{figure}

\subsection{Incremental Attribution Setting}
In realistic scenarios, new generative models are released continuously. We therefore reformulate attribution as a class-incremental learning process. The continual evolution of models is represented as a temporal sequence of tasks
\begin{equation}
\mathcal{T}_0, \mathcal{T}_1, \dots, \mathcal{T}_t, \quad \mathcal{T}_i = \{(\mathbf{x}, y) \mid y \in \mathcal{Y}_i\},
\label{eq:task_sequence}
\end{equation}
where \(\mathcal{Y}_i\) is the label set of new generative models introduced in task \(\mathcal{T}_i\), and \(\mathcal{Y}_i \cap \mathcal{Y}_j = \emptyset\) for \(i \neq j\). The cumulative label space at step \(t\) is \(\mathcal{Y}_{\le t} = \bigcup_{i=0}^{t} \mathcal{Y}_i\).

At each step, the attribution model \(F\) learns on \(\mathcal{T}_t\) while retaining performance on all previous tasks. The learning objective is   
\begin{equation}
\min \sum_{i=0}^{t}\mathbb{E}_{(x,y)\sim \mathcal{T}_i}\!\left[\ell\!\left(F\big(x)\big),y\right)\right],
\label{eq:incremental_objective}
\end{equation}
where \(\ell\) denotes the classification loss. This formulation explicitly models the dynamic expansion of the attribution label space and serves as the foundation for our hierarchical and replay-based approach introduced in Section~\ref{sec:method}.

\subsection{Incremental Attribution Benchmark}
To evaluate this setting, we construct the \textbf{Incremental Attribution Benchmark (IABench)}, which covers 28 generative models released between 2022 and 2025, including GAN, diffusion, and autoregressive families. Each model represents a distinct category, and data are divided into temporally ordered training and testing splits to emulate the evolution of real generative models. IABench provides a realistic and systematic foundation for benchmarking attribution methods under continual and open-set conditions. The IABench dataset includes 4 categories of GANs \cite{e4s2023, r3gan2024, stylegan32021, styleganxl2022}, 2 categories of autoregressive models \cite{gpt4o2025, cogview2021}, and 22 categories of diffusion models \cite{sdxl2024,lcm2023,ldm2022,tinysd, flux12024, pa2024,pg2024, midjourney2023, imagen32024, nanobanana2025,ssd2024,dalle32023}. To better demonstrate the model's attribution capabilities while more closely aligning with real-world application scenarios, we incorporate models with closer version numbers in IABench, such as Stable Diffusion 1.4, 1.5, and 2.0. This enables a more comprehensive evaluation of attribution methods' ability to discriminate subtle artifact differences between proximate versions of the same generative model. We will detail IABench in Appendix(\cref{sec:sup_dataset}).

\begin{figure*}[ht]
    \centerline{\includegraphics[width=0.95\textwidth]{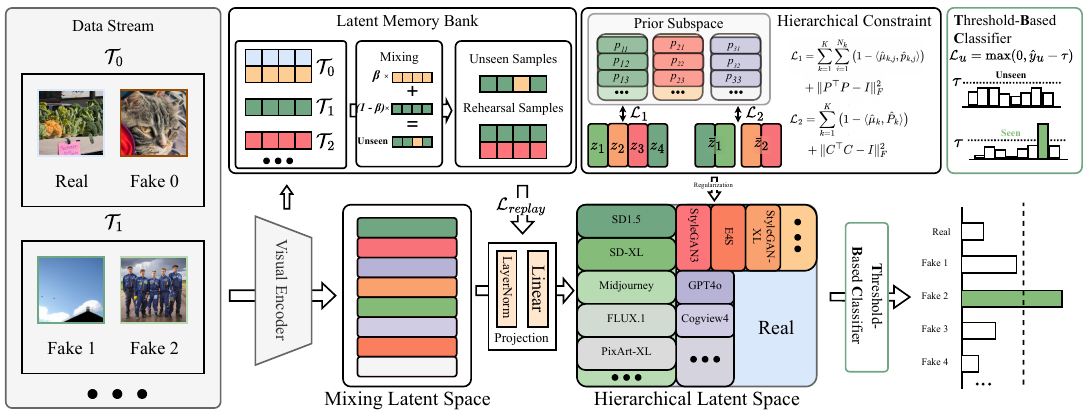}}
    \vspace{-0.05in}
    \caption{\textbf{Pipeline of IncreFA.} IncreFA consists of two components: \textbf{Hierarchical Constraints} and \textbf{Latent Memory Bank}.  For tasks in the data stream, visual features are first extracted using a pretrained large model, from which samples are drawn to form the latent memory bank. Subsequently, the visual features are projected to obtain a feature space with a hierarchical structure, on which classification is performed. The latent memory bank not only handles replay for old tasks but also employs \textbf{Random Mixing} to simulate unseen samples, thereby enhancing the model's detection capability for unseen instances.}
    \vspace{-0.15in}
    \label{fig:pipeline}
\end{figure*}

\section{Method}
\label{sec:method}

We propose \textbf{IncreFA}, a continual and open-set attribution framework that exploits the hierarchical lineage of generative models. As shown in Figure~\ref{fig:pipeline}, IncreFA operates on a frozen pretrained encoder to extract stable visual features and learns two complementary mechanisms: (\textit{i}) hierarchical constraints that structure the latent space according to architectural ancestry, and (\textit{ii}) a latent memory bank that preserves compact exemplars for continual replay and unseen simulation. Together, these components enable the model to adapt to newly released generators without catastrophic forgetting.

\subsection{Feature Representation and Projection}
Let \(\mathbf{f}(x)\in\mathbb{R}^d\) denote the feature extracted from a frozen encoder such as CLIP. The encoder provides semantically consistent representations across different generative models. A lightweight projection head \(G(\cdot)\) maps these features into the structured latent space \(\mathcal{Z}\) defined in Eq.~\ref{eq:hierarchy}:
\begin{equation}
z = G(\mathbf{f}(x)), \quad z \in \mathcal{Z}.
\label{eq:projection}
\end{equation}
The final attribution classifier \(F(\cdot)\) maps latent vectors \(z\) to class logits and is used for both attribution and open-set recognition.

\subsection{Hierarchical Constraint Loss}
To align latent representations with the hierarchical structure of generative models, IncreFA introduces two complementary constraints: fine-grained orthogonality and coarse-level regularisation. 
Both levels of anchors, \(\hat{p}_{k,j}\) for models and \(\hat{P}_k\) for families, are jointly learned, forming a hierarchically coupled prior system that captures both family-shared and model-specific semantics.
\vspace{-0.1in}
\paragraph{Fine-Grained Orthogonality.}
For each model \(\mathcal{M}_{k,j}\) within family \(k\), we define a unit-norm prototype
\begin{equation}
\hat{\mu}_{k,j} = \mathrm{Normalize}\left(\mathbb{E}_{x\sim\mathcal{M}_{k,j}}[\,G(\mathbf{f}(x))\,]\right).
\end{equation}
Each prototype is encouraged to align with its corresponding learnable orthogonal anchor \(\hat{p}_{k,j}\) while remaining distinct from others:
\begin{equation}
\vspace{-0.02in}
\mathcal{L}_1 = \sum_{k=1}^{K}\sum_{j=1}^{N_k}\big(1 - \langle \hat{\mu}_{k,j}, \hat{p}_{k,j} \rangle\big) + \|Q^\top Q - I\|_F^2,
\label{eq:l1}
\vspace{-0.05in}
\end{equation}
where \(Q=[\hat{p}_{1,1},\dots,\hat{p}_{K,N_K}]\) and $I$ denotes the identity matrix. The first term promotes angular alignment, while the orthogonality term regularises the anchors to remain mutually orthogonal, ensuring angular separation among all models.
\vspace{-0.1in}
\paragraph{Coarse-Level Regularisation.}
At the family level, we compute the mean prototype
\begin{equation}
\hat{\mu}_k = \frac{1}{N_k}\sum_{j=1}^{N_k}\hat{\mu}_{k,j},
\end{equation}
and align it with a family-level anchor \(\hat{P}_k\):
\begin{equation}
\mathcal{L}_2 = \sum_{k=1}^{K}\big(1 - \langle \hat{\mu}_k, \hat{P}_k \rangle\big) + \|C^\top C - I\|_F^2,
\label{eq:l2}
\end{equation}
where \(C=[\hat{P}_1,\dots,\hat{P}_K]\). This regularisation captures architectural invariants within each family while maintaining orthogonality across different families. Together, \(\mathcal{L}_1\) and \(\mathcal{L}_2\) implement the hierarchical decomposition of Eq.~\ref{eq:hierarchy}, promoting intra-family coherence and inter-family disentanglement.

\subsection{Latent Memory Bank}
To enable continual adaptation, IncreFA maintains a \textbf{Latent Memory Bank} that stores compact encoder features rather than raw images, preserving memory efficiency and preventing data leakage. At incremental step \(t\), the bank \(\mathcal{B}_{t-1}\) contains encoder features \(\mathbf{f}(x)\) from previous tasks \(\{\mathcal{T}_0, \dots, \mathcal{T}_{t-1}\}\). During the training of \(\mathcal{T}_t\), replayed samples are used to optimize  
\begin{equation}
\mathcal{L}_{replay} = \mathbb{E}_{(\mathbf{f},y)\sim \mathcal{B}_{t-1}}\big[\ell(F(G(\mathbf{f})), y)\big],
\label{eq:replay}
\end{equation}
where \(\ell\) denotes the cross-entropy loss. Features are re-encoded through the current \(G\) to avoid projection drift across tasks. Each class retains a fixed budget of exemplars selected by herding, balancing performance and memory usage. The latent-level representation reduces memory consumption by over two orders of magnitude compared with raw-image replay, enabling data-efficient continual learning.
\vspace{-0.05in}
\paragraph{Random Mixing for Unseen Simulation.}
To enhance robustness against unseen generators, IncreFA creates pseudo-unseen samples by randomly interpolating latent pairs from different classes:
\begin{equation}
z_u = \beta z_1 + (1 - \beta) z_2, \quad \beta \sim \text{U}(0,1),
\label{eq:mixing}
\end{equation}
where \(z_1, z_2 = G(\mathbf{f}_1), G(\mathbf{f}_2)\) are sampled from distinct classes. The resulting \(z_u\) typically lies in low-density regions near decision boundaries, encouraging the model to identify transition zones between known classes and improving open-set calibration.

\subsection{Open-Set Recognition}
For open-set attribution, we constrain the classifier’s confidence on pseudo-unseen samples. Given classifier output \(\hat{y}_u = F(z_u)\), we penalise excessive confidence as
\begin{equation}
\mathcal{L}_u = \max(0, \max(\text{softmax}(\hat{y}_u)) - \tau),
\label{eq:lu}
\end{equation}
where \(\tau\) is a confidence threshold selected on a held-out calibration split after each task.  At inference, a test image \(x\) is considered unseen when
\begin{equation}
\max(\text{softmax}(F(G(\mathbf{f}(x))))) < \tau,
\label{eq:openset}
\end{equation}
ensuring consistent open-set behaviour as the number of known models evolves. The calibrated threshold \(\tau\) balances false alarms and missed detections, ensuring reliable detection of unseen generators under a continually expanding model set.

\subsection{Classification Loss}
For each incremental task \(\mathcal{T}_t\), the model performs standard supervised learning on the current dataset \(\mathcal{D}_t = \{(x,y)\}\). 
The classification loss is defined as the expectation of cross-entropy between the predicted logits and ground-truth labels:
\begin{equation}
\mathcal{L}_{cls} = \mathbb{E}_{(x,y)\sim\mathcal{D}_t}\big[\ell(F(G(\mathbf{f}(x))), y)\big],
\label{eq:lcls}
\end{equation}
where \(\ell(\cdot)\) the cross-entropy loss. 
This term serves as the primary discriminative objective for model attribution at each incremental step.

\subsection{Overall Objective}
The final objective unifies classification, hierarchical regularisation, continual replay, and open-set calibration:
\begin{equation}
\mathcal{L} = \mathcal{L}_{cls} + \alpha_1\mathcal{L}_1 + \alpha_2\mathcal{L}_2 + \alpha_3\mathcal{L}_u + \alpha_4\mathcal{L}_{replay}.
\label{eq:final}
\end{equation}
This formulation consolidates the hierarchical structure in Eq.~\ref{eq:hierarchy} and the incremental objective in Eq.~\ref{eq:incremental_objective}, establishing a principled foundation for scalable, continual, and open-set image attribution.

\begin{table*}[!t]
\centering
\caption{\textbf{Comparative Experiments with Other Incremental Learning Methods.} Following \textbf{EP1}, in our experiments, we report the average accuracy for the current task, and upon completion of the final task, we report the unseen detection rate. \note We designate the more recently released Nano-banana and Imagen3 as unseen models to evaluate the model's detection capability for unseen instances. \nnote We directly employ a frozen CLIP model with a linear layer for training.}
\vspace{-0.1in}
\resizebox{0.9\linewidth}{!}{
\begin{tabular}{ll|cccccccc|c@{ }c}
\toprule
Method & Reference & $\mathcal{T}_0$ & $\mathcal{T}_1$ & $\mathcal{T}_2$ & $\mathcal{T}_3$ & $\mathcal{T}_4$ & $\mathcal{T}_5$ & $\mathcal{T}_6$ & $\mathcal{T}_7$ & Un. Acc.\note & Auth. Acc. \\
\midrule
Vanilla baseline\nnote & & 99.65 & 61.68 & 50.31 & 48.89 & 44.41 & 40.39 & 35.29 & 35.24 & 69.13 & 85.92 \\
ICaRL \cite{icarl2017} & \emph{CVPR'2017} & 98.12 & \underline{93.93} & 80.98 & \underline{83.23} & 74.66 & 71.99 & 72.00 & 72.31 & 83.00 & 84.81 \\
Foster \cite{foster2022} & \emph{ECCV'2022} & 99.87 & 92.89 & 78.27 & 76.35 & 72.10 & 72.86 & 69.92 & 71.39 & 72.33 & 93.18 \\
DualPrompt \cite{dualprompt2022} & \emph{ECCV'2022} & \textbf{99.98} & 82.32 & 75.49 & 50.41 & 50.26 & 46.83 & 40.30 & 35.12 & 62.12 & 85.62\\
L2P \cite{l2p2022} & \emph{CVPR'2022} & 89.17 & 59.43 & 45.99 & 40.20 & 29.42 & 23.26 & 29.88 & 29.51 & 79.51 & 86.26\\
SimpleCIL \cite{simplecil2025} & \emph{IJCV'2024} & 96.74 & 75.40 & 56.48 & 47.14 & 41.69 & 39.58 & 34.68 & 31.68 & 79.13 & 95.50\\
APER-Aperpter \cite{simplecil2025}& \emph{IJCV'2024} & 99.22 & 76.28 & 68.70 & 61.36 & 51.15 & 46.71 & 45.82 & 41.51 & 85.00 & 98.04\\
DGR \cite{dgr2024} & \emph{CVPR'2024} & 98.36 & 93.41 & \underline{84.28} & 83.14 & \underline{75.46} & \underline{76.99} & \underline{77.86} & \underline{75.68} & \underline{94.21} & 97.18\\
TUNA \cite{tuna2025} & \emph{CVPR'2025} & 99.41 & 87.84 & 77.33 & 77.19 & 75.31 & 71.58 & 64.18 & 63.87 & 92.10 & 93.79 \\
MOS \cite{mos2025} & \emph{AAAI'2025} & 99.93 & 92.75 & 81.83 & 79.55 & 75.05 & 68.44 & 68.61 & 66.84 & 91.92 & 98.58\\
\rowcolor{background_gray} IncreFA & Ours & \textbf{99.99} &\textbf{ 99.17} & \textbf{95.91} & \textbf{92.94} & \textbf{88.09} & \textbf{80.69} & \textbf{79.17} & \textbf{78.80} & \textbf{98.93} & \textbf{99.97}\\
\bottomrule
\end{tabular}}
\vspace{-0.15in}
\label{tab:incre_compare}
\end{table*}

\section{Experiments}

\subsection{Settings}
\label{sec:settings}

To comprehensively evaluate the continual, hierarchical, and open-set properties of IncreFA, we design two complementary protocols: (\textbf{EP1}) Incremental Attribution, and (\textbf{EP2})Static Attribution.
\vspace{-0.1in}
\paragraph{Evaluation Protocol 1 (EP1): Incremental Attribution.}
EP1 simulates a realistic generative landscape in which new models appear sequentially.  
Each task \(\mathcal{T}_t\) introduces \(L=4\) new generative models, reflecting the release pace observed between 2022–2025 in IABench.  
The initial task \(\mathcal{T}_0\) contains the real-image class and one randomly selected generator.  
After training for \(n\) epochs, the model receives the next task \(\mathcal{T}_{t+1}\) without access to previous raw data.  
To emulate open-world conditions, the most recently released models (e.g., Nano-Banana and Imagen3) are withheld as \emph{unseen generators} and used exclusively for open-set evaluation.

At the end of each task, we compute:  
(1) the current average attribution accuracy across all seen generators; and  
(2) unseen detection accuracy based on the open-set criterion in Eq.~\ref{eq:openset}.  
Since several baselines lack explicit open-set modules, only compatible methods are reported under this metric.  
This protocol directly evaluates IncreFA’s continual learning and open-set generalisation ability.
\vspace{-0.1in}

\paragraph{Evaluation Protocol 2 (EP2): Static Attribution.}
 EP2 serves as a static closed-world evaluation to isolate attribution performance from continual effects.  
All training data across 28 generators are jointly used without task splits.  
The objective is to assess IncreFA’s attribution and authenticity recognition capacity under static conditions, providing a controlled comparison with conventional classifier-based baselines.  
\vspace{-0.1in}
\paragraph{Metrics:}  
(1) \textit{Average Accuracy (\textbf{Avg. Acc.})} for multi-class attribution across all generators;  
(2) \textit{Authenticity Accuracy (\textbf{Auth. Acc.})} for distinguishing real versus generated images.  
(3) \textit{Unseen Accuracy (\textbf{Un. Acc.})} for identifying unseen images.
EP2 complements EP1 by validating IncreFA’s upper-bound performance under fully supervised training.

\begin{table}[!t]
\caption{\textbf{Comparison with Other Attribution Methods.} The experimental setup employs EP2, with metrics comprising per-model attribution accuracy and final average accuracy. \note We designate the more recently released Nano-banana and Imagen3 as unseen models to evaluate the model's detection capability for unseen instances. \nnote We directly employ a frozen CLIP model with a linear layer for training.}
\small
\label{tab:attr_compare}
\centering
\resizebox{0.98\linewidth}{!}{
\begin{tabular}{@{ }c|@{  }l@{ }c@{ }c@{ }c@{ }c@{ }c@{ }c@{ }c@{}}
\toprule
                          & & Vanilla\nnote & DNA-Net  & RepMix    & DE-FAKE  & POSE      & Siamese         & Ours \\
\midrule
 & Real & 97.44 & 95.88 & \underline{99.52} & 93.23 & 97.92 & 94.43 & \textbf{99.66} \\
\midrule
\multirow{4}{*}{\rotatebox{90}{GANs}} & StyleGAN-XL & 98.82 & 91.40 &\textbf{ 100.00} & 79.12 & 88.29 & 98.43 & \underline{99.46} \\
 & StyleGAN3 & \underline{99.83} & 91.00 & \textbf{100.00} & 89.00 & 69.50 & 91.98 & 99.77 \\
 & E4S & 30.90 & \textbf{100.00} & \textbf{100.00} & 97.09 & \underline{99.43} & 98.79 & 94.60 \\
 & R3GAN & 98.40 & 45.80 & 74.63 & 88.67 & 98.23 & \textbf{99.90} & \underline{99.56} \\
\midrule
\multirow{22}{*}{\rotatebox{90}{Diffusions}} & LDM & 97.00 & 98.80 & \underline{99.34} & 88.81 & 87.38 & 87.00 & \textbf{99.84} \\
 & SD V1.4 & \underline{97.38} & 71.17 & 81.82 & 92.19 & 77.21 & 92.36 & \underline{99.56} \\
 & SD V1.5 & \underline{87.12} & 21.00 & 17.56 & 15.51 & 86.46 & 81.02 & \underline{98.42} \\
 & SSD-1B & 83.32 & 76.40 & \textbf{93.90} & 79.20 & 66.05 & 89.10 & \underline{89.04} \\
 & Tiny-SD & 93.90 & 65.20 & 90.74 & \textbf{98.09} & 76.15 & 79.31 & \underline{96.86} \\
 & SegMoE-SD & \underline{99.13} & 36.20 & 88.27 & 96.54 & 76.91 & 92.93 & \textbf{99.89} \\
 & Small-SD & \underline{98.04} & 85.40 & 85.40 & 81.96 & 77.26 & 94.52 & \textbf{98.72} \\
 & SD V2.1 & \underline{94.76} & 13.00 & 80.48 & 45.81 & 19.87 & 84.30 & \textbf{95.96} \\
 & SD V3 & 15.92 & 77.60 & 77.43 & \underline{99.70} & 30.49 & 89.75 & \textbf{100.00} \\
 & SDXL-turbo & 98.92 & 99.20 & \textbf{99.84} & 47.57 & 89.53 & 59.93 & \underline{99.28} \\
 & SD V2 & 19.02 & 21.00 & 74.76 & 71.27 & 39.91 & \underline{94.50} & \textbf{97.21} \\
 & SD XL & 74.50 & 14.60 & \textbf{97.18} & 32.90 & 70.51 & \underline{86.48} & 80.00 \\
 & PG V2.5 & 85.90 & 96.00 & \textbf{99.62} & 39.30 & 41.59 & 65.14 & \underline{97.40} \\
 & PG V2 & 99.74 & 3.00 & \underline{99.60} & 77.23 & 86.58 & 82.22 & \textbf{99.71} \\
 & PAXL V2 & 85.90 & 24.80 & \textbf{97.70} & 76.50 & 77.11 & 79.41 & \underline{91.78} \\
 & LCM-SD XL & 96.70 & 80.60 & \textbf{99.10} & 96.23 & 86.88 & 59.99 & \underline{98.48} \\
 & LCM-SD V1.5 & 92.64 & 50.20 & \underline{98.52} & 95.32 & 86.46 & \textbf{99.21} & 95.90 \\
 & FLUX.1-sch & \underline{98.90} & 29.60 & \textbf{99.50} & 92.10 & 54.50 & 85.59 & 72.66 \\
 & FLUX.1-dev & 75.14 & 55.60 & \textbf{99.32} & 35.92 & 50.86 & 19.82 & \underline{98.14} \\
 & Midjourney-V6 & 91.36 & 16.80 & \textbf{95.00} & 39.99 & 52.28 & 66.02 & \underline{93.12} \\
\midrule
\multirow{2}{*}{\rotatebox{90}{ARs}} & GPT-4o & 95.56 & \underline{97.00} & 95.00 & 86.65 & 79.97 & 95.98 & \textbf{97.54} \\
 & Cogview4 & 98.56 & 51.80 & \textbf{99.97} & 92.67 & 89.64 & 88.34 & \underline{99.12} \\
\midrule
 & Avg. Acc. & 83.75 & 71.61 & \underline{91.16} & 73.97 & 74.69 & 83.57 & \textbf{95.93} \\
 & Auth. Acc. & 97.66 & \underline{95.99} & 95.65 & 90.90 & 84.22 & 89.79 & \textbf{99.97} \\
 & Un. Acc.\note & 70.04 & 78.40 & 82.13 & 72.23 & \underline{98.23} & 96.23 & \textbf{98.78} \\
\bottomrule
\end{tabular}}
\vspace{-0.2in}
\end{table}

\subsection{Implementation Details}
We adopt ViT-B/16 as the frozen visual backbone and update only the projection head \(G(\cdot)\) and classifier \(F(\cdot)\). 
The backbone is initialised from CLIP and kept fixed throughout all incremental tasks to ensure stable feature semantics across generators. 
Training is performed using the Adam optimiser with a learning rate of 1e-3. 
The loss coefficients are set as \(\alpha_1=0.2\), \(\alpha_2=0.5\), \(\alpha_3=0.5\), and \(\alpha_4=1.0\) for all experiments and $\tau$ is set to 0.65. 
Each task \(\mathcal{T}_t\) is trained for four epochs with a batch size of 512 with $L=4$.
For latent replay, 150 exemplars per class are retained using herding selection to maintain diversity in latent space. 
All experiments are conducted on a Nvidia L20 GPU  and all images are resized to \(256\times256\). 
Following Eq.~\ref{eq:mixing}, pseudo-unseen samples are mixed at a ratio \(\beta\in[0,1]\) drawn uniformly for each mini-batch.

\subsection{Comparisons Under EP1}

We evaluate IncreFA under the incremental attribution protocol (EP1) against both classical and recent class-incremental approaches, all using the same frozen ViT-B/16 backbone for fairness. Baselines include rehearsal-based methods (ICaRL \cite{icarl2017}, DGR \cite{dgr2024}, MOS \cite{mos2025}), prompt-based methods (L2P \cite{l2p2022}, DualPrompt \cite{dualprompt2022}, TUNA \cite{tuna2025}), and regularisation- or representation-based variants (Foster \cite{foster2022}, APER \cite{simplecil2025}, SimpleCIL \cite{simplecil2025}), with naive fine-tuning as a reference. We will detail baselines in Appendix(\cref{sec:sup_exp}).

Table~\ref{tab:incre_compare} shows that all incremental methods exhibit progressive degradation as new generators are introduced, reflecting their difficulty in balancing plasticity and retention across overlapping distributions. 
Prompt-based methods such as L2P \cite{l2p2022} and DualPrompt \cite{dualprompt2022} degrade sharply in this setting. Their semantic prompts fail to capture generator-specific artefacts, leading to collapsed prompt diversity and severe forgetting. 
On the contrary, 
IncreFA maintains stable attribution throughout the entire sequence, achieving an attribution accuracy of 78.80\% even in the final task and surpassing the second-place method by 3.12\%. Furthermore, IncreFA maintains excellent unseen detection efficiency, achieving 98.93\%.

\subsection{Comparisons Under EP2}

We further evaluate IncreFA under the closed attribution protocol (EP2) to assess its upper-bound attribution capability without incremental constraints. All methods use the same ViT-B/16 backbone and are trained jointly on all generators. We compare with representative approaches including CLIP \cite{clip2021}, DNA-Net \cite{dna2022}, RepMix \cite{repmix2022}, DE-FAKE \cite{defake2023}, POSE \cite{pose2023}, and Siamese \cite{siamese2024}. We will detail baselines in Appendix(\cref{sec:sup_exp}).

As shown in Table~\ref{tab:attr_compare}, IncreFA achieves the highest average attribution accuracy of \textbf{95.93\%} and a near-perfect detection rate of \textbf{99.97\%}. It consistently outperforms CLIP and RepMix, with larger gains on recent diffusion models such as SD-XL, SD3, and FLUX, where other methods struggle.
These results verify that the hierarchical representation learned by IncreFA improves not only continual adaptation but also static discrimination. By aligning latent features within structured subspaces, IncreFA captures subtle inter-generator artifacts that traditional feature-based approaches often overlook. The model thus provides a unified solution that remains accurate and robust across both incremental and closed-world attribution settings.

\subsection{Ablation Study}

\begin{table}[!t]
\centering
\caption{\textbf{Ablation on Core Components of IncreFA.}}
\vspace{-0.1in}
\label{tab:main_abalation}
\resizebox{0.95\linewidth}{!}{
\begin{tabular}{l|cccccccc|c}
\toprule
Components        & $\mathcal{T}_0$  & $\mathcal{T}_1$  & $\mathcal{T}_2$  & $\mathcal{T}_3$  & $\mathcal{T}_4$  & $\mathcal{T}_5$  & $\mathcal{T}_6$  & $\mathcal{T}_7$   & Unseen \\ 
\midrule
baseline                &   99.65  & 61.68  &  50.31  & 48.89  &  44.41 & 40.39 & 35.29  & 35.24  & 69.13 \\
+ $\mathcal{L}_{replay}$&   99.18  &  84.08 &  80.18  & 68.14  &  67.42 & 61.08 & 58.92  & 52.99  & 73.98 \\
+ $\mathcal{L}_1$       &   99.05  &  84.19 &  81.94  & 79.86  &  83.24 & 71.14 & 72.35  & 73.16  & 72.19 \\
+ $\mathcal{L}_2$       &   99.81  &  98.94 &  94.94  & \textbf{92.99}  &  \textbf{89.24} & 79.14 & \textbf{79.35}  & 76.16  & 81.94 \\
+ $\mathcal{L}_{u}$     &   \textbf{99.99}  &  \textbf{99.17} &  \textbf{95.91}  & 92.94  & 88.09  & \textbf{80.69} &  79.17 & \textbf{78.80}  & \textbf{98.93} \\
\bottomrule
\end{tabular}}
\vspace{-0.15in}
\end{table}

\paragraph{Ablation on Components .}  
We ablate the four key components of IncreFA on the Vanilla baseline (\cref{tab:main_abalation}) to examine how each contributes to continual stability and generalisation.  
Introducing \(\mathcal{L}_{replay}\) alleviates the sharp degradation observed in early increments, confirming that latent replay effectively preserves short-term memory.  
Building upon this, the hierarchical constraints \(\mathcal{L}_{1}\) and \(\mathcal{L}_{2}\) further restructure the latent geometry: \(\mathcal{L}_{1}\) enhances model-level separability, while \(\mathcal{L}_{2}\) captures family-level invariants, together ensuring consistent retention as the task sequence grows.  
Finally, integrating the open-set regularisation term \(\mathcal{L}_{u}\) calibrates the decision boundary and markedly improves unseen detection, extending IncreFA’s robustness beyond the closed set.  
Overall, these results reveal a clear progression that replay stabilises memory, hierarchical constraints preserve structure, and open-set calibration enables reliable generalisation in evolving generative landscapes.

\vspace{-0.15in}
\paragraph{Ablation on Hyperparameters.}  
We evaluate the impact of $\alpha_1$–$\alpha_4$, which balance hierarchical regularisation, open-set calibration, and replay.  
As shown in Figure~\ref{tab:ablation_hyper}, the left figure examines $\alpha_1$ and $\alpha_2$—controlling fine-grained and coarse-grained hierarchical constraints, while the right figure analyses $\alpha_3$ and $\alpha_4$, corresponding to open-set and replay terms.  

In the left figure, $\alpha_1$ enforces model-level orthogonality and $\alpha_2$ enhances family-level consistency.  
Moderate values (\(0.2\!\sim\!0.5\)) yield the best balance between structural disentanglement and adaptability; excessive $\alpha_1$ causes over-regularisation and degraded generalisation.  
In the right figure, $\alpha_4$ strongly governs continual stability, while $\alpha_3$ has a minor but steady effect on calibration.  
Overall, IncreFA is robust across a broad range, with performance mainly driven by the synergy between hierarchical constraints and latent replay.  
Due to space constraints, unseen detection results are provided in the supplementary materials (\cref{sec:sup_exp}).

\begin{figure}[!t]
    \centerline{\includegraphics[width=\columnwidth]{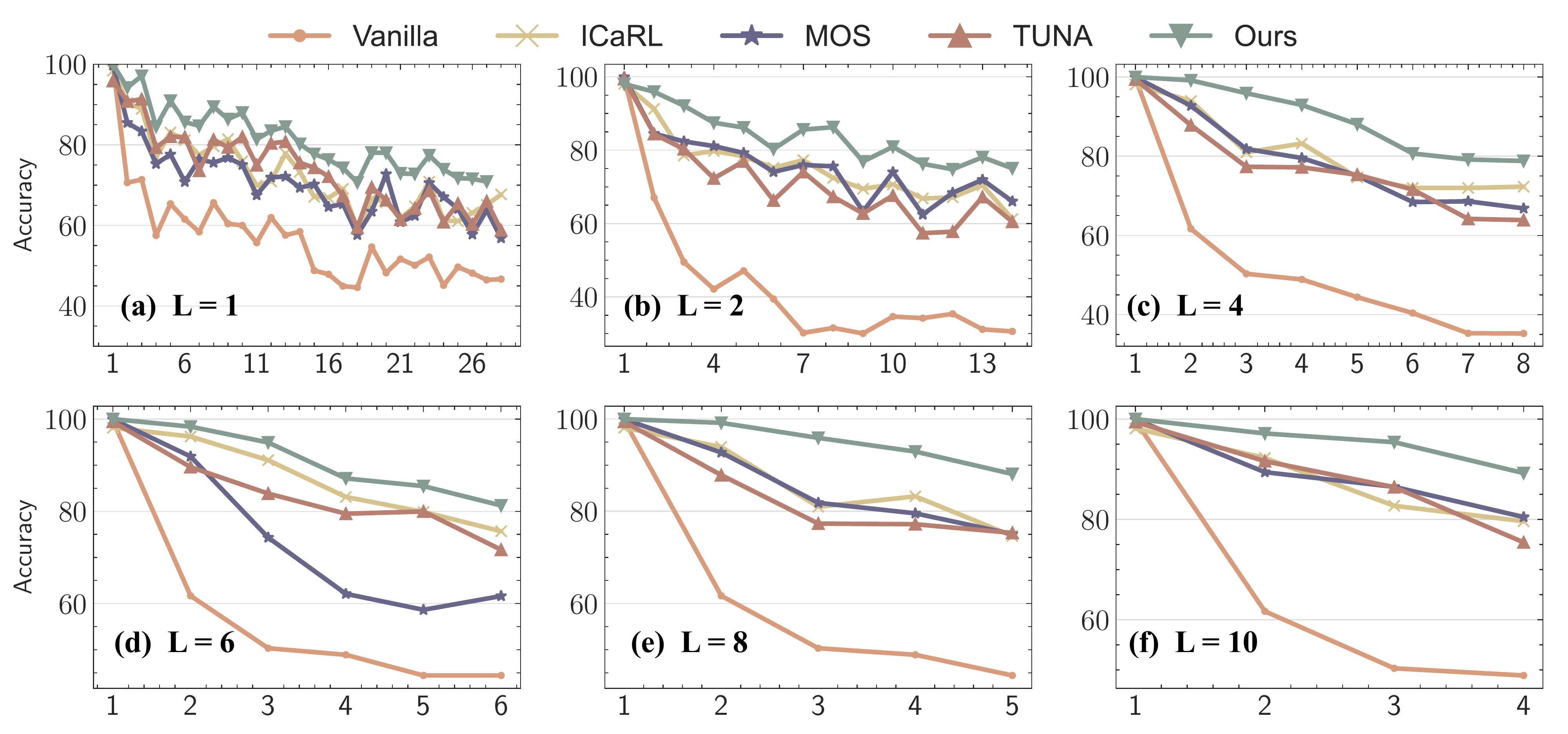}}
    \vspace{-0.1in}
    \caption{Impact of different \( L \) on various incremental methods.}
    \vspace{-0.25in}
    \label{fig:line-vis}
\end{figure}

\begin{table}[!t]
\centering
\caption{\textbf{Ablations on hyperparameters.} \textbf{Left:} Ablation on $\alpha_1$ and $\alpha_2$. \textbf{Right:} Ablation on $\alpha_3$ and $\alpha_4$. Note that for left, we fix \(\alpha_3\) and \(\alpha_4\) to 0.5 and 1.0. For right, we fix \(\alpha_1\) and \(\alpha_2\) to 0.2 and 0.5, respectively.}
\begin{minipage}[!t]{0.49\columnwidth}
\centering
\vspace{-0.1in}
\resizebox{\linewidth}{!}{
\begin{tabular}{c|cccccc}
\toprule
\diagbox{$\alpha_1$}{$\alpha_2$} & 0.0 & 0.1 & 0.2 & 0.5 & 0.8 & 1.0 \\
\midrule
0.0 & \cellcolor{cell1_r0_c0}54.03 & \cellcolor{cell1_r0_c1}59.20 & \cellcolor{cell1_r0_c2}62.95 & \cellcolor{cell1_r0_c3}67.06 & \cellcolor{cell1_r0_c4}73.76 & \cellcolor{cell1_r0_c5}70.11\\
0.1 & \cellcolor{cell1_r1_c0}65.90 & \cellcolor{cell1_r1_c1}67.29 & \cellcolor{cell1_r1_c2}69.35 & \cellcolor{cell1_r1_c3}69.39 & \cellcolor{cell1_r1_c4}66.01 & \cellcolor{cell1_r1_c5}65.26\\
0.2 & \cellcolor{cell1_r2_c0}64.71 & \cellcolor{cell1_r2_c1}71.92 & \cellcolor{cell1_r2_c2}78.19 & \cellcolor{cell1_r2_c3}\textbf{78.32} & \cellcolor{cell1_r2_c4}77.51 & \cellcolor{cell1_r2_c5}69.77\\
0.5 & \cellcolor{cell1_r3_c0}67.51 & \cellcolor{cell1_r3_c1}66.00 & \cellcolor{cell1_r3_c2}65.67 & \cellcolor{cell1_r3_c3}68.18 & \cellcolor{cell1_r3_c4}71.94 & \cellcolor{cell1_r3_c5}70.99\\
0.8 & \cellcolor{cell1_r4_c0}66.48 & \cellcolor{cell1_r4_c1}68.98 & \cellcolor{cell1_r4_c2}72.00 & \cellcolor{cell1_r4_c3}71.82 & \cellcolor{cell1_r4_c4}71.09 & \cellcolor{cell1_r4_c5}70.31\\
1.0 & \cellcolor{cell1_r5_c0}66.59 & \cellcolor{cell1_r5_c1}67.99 & \cellcolor{cell1_r5_c2}71.00 & \cellcolor{cell1_r5_c3}71.12 & \cellcolor{cell1_r5_c4}69.39 & \cellcolor{cell1_r5_c5}63.44\\
\bottomrule
\end{tabular}
}
\label{tab:ablation_weight_1}
\end{minipage}
\hfill
\begin{minipage}[!t]{0.49\columnwidth}
\centering
\vspace{-0.1in}
\resizebox{\linewidth}{!}{
\begin{tabular}{c|cccccc}
\toprule
\diagbox{$\alpha_3$}{$\alpha_4$} & 0.0 & 0.1 & 0.2 & 0.5 & 0.8 & 1.0 \\
\midrule
0.0 & \cellcolor{cell2_r0_c0}37.85 & \cellcolor{cell2_r0_c1}49.19 & \cellcolor{cell2_r0_c2}55.15 & \cellcolor{cell2_r0_c3}69.91 & \cellcolor{cell2_r0_c4}70.86 & \cellcolor{cell2_r0_c5}72.59\\
0.1 & \cellcolor{cell2_r1_c0}34.13 & \cellcolor{cell2_r1_c1}48.00 & \cellcolor{cell2_r1_c2}51.54 & \cellcolor{cell2_r1_c3}54.16 & \cellcolor{cell2_r1_c4}57.14 & \cellcolor{cell2_r1_c5}65.26\\
0.2 & \cellcolor{cell2_r2_c0}21.00 & \cellcolor{cell2_r2_c1}50.21 & \cellcolor{cell2_r2_c2}67.55 & \cellcolor{cell2_r2_c3}59.34 & \cellcolor{cell2_r2_c4}67.00 & \cellcolor{cell2_r2_c5}73.77\\
0.5 & \cellcolor{cell2_r3_c0}35.01 & \cellcolor{cell2_r3_c1}54.01 & \cellcolor{cell2_r3_c2}60.96 & \cellcolor{cell2_r3_c3}75.60 & \cellcolor{cell2_r3_c4}77.99 & \cellcolor{cell2_r3_c5}\textbf{78.80}\\
0.8 & \cellcolor{cell2_r4_c0}33.94 & \cellcolor{cell2_r4_c1}41.59 & \cellcolor{cell2_r4_c2}42.30 & \cellcolor{cell2_r4_c3}61.98 & \cellcolor{cell2_r4_c4}62.33 & \cellcolor{cell2_r4_c5}77.31\\
1.0 & \cellcolor{cell2_r5_c0}29.18 & \cellcolor{cell2_r5_c1}10.56 & \cellcolor{cell2_r5_c2}41.39 & \cellcolor{cell2_r5_c3}57.23 & \cellcolor{cell2_r5_c4}65.10 & \cellcolor{cell2_r5_c5}74.44\\
\bottomrule
\end{tabular}
}
\label{tab:ablation_weight_2}
\end{minipage}
\vspace{-0.1in}
\label{tab:ablation_hyper}
\end{table}
\vspace{-0.1in}
\paragraph{Ablation on $\mathbf{L}$ .}  
We further study the impact of task granularity by varying \(L\), the number of new generative models introduced per task \(\mathcal{T}\).  
Each setting controls the dataset composition \(\mathcal{D}\) within successive tasks and compares different incremental methods under identical conditions.  
As shown in \cref{fig:line-vis}, IncreFA consistently achieves state-of-the-art performance across all values of \(L\).  
The slight fluctuations in incremental accuracy arise from the varying attribution difficulty of different \(\mathcal{D}\) across tasks.  
In all cases, the initial task \(\mathcal{T}_0\) includes real images and one randomly selected generative model, after which increments proceed according to the specified \(L\), and the average accuracy at each stage is reported.

\subsection{Visualizations}

\begin{figure}[!t]
    \centerline{\includegraphics[width=\columnwidth]{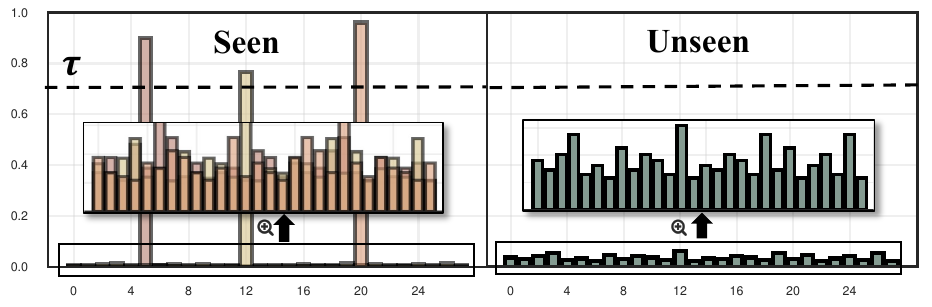}}
    \vspace{-0.1in}
    \caption{\textbf{Visualizations between Seen and Unseen.} We have zoomed in to showcase the subtle distribution differences.}
    \vspace{-0.1in}
    \label{fig:logits-vis}
\end{figure}

\begin{figure}[!t]
    \centerline{\includegraphics[width=1.0\columnwidth]{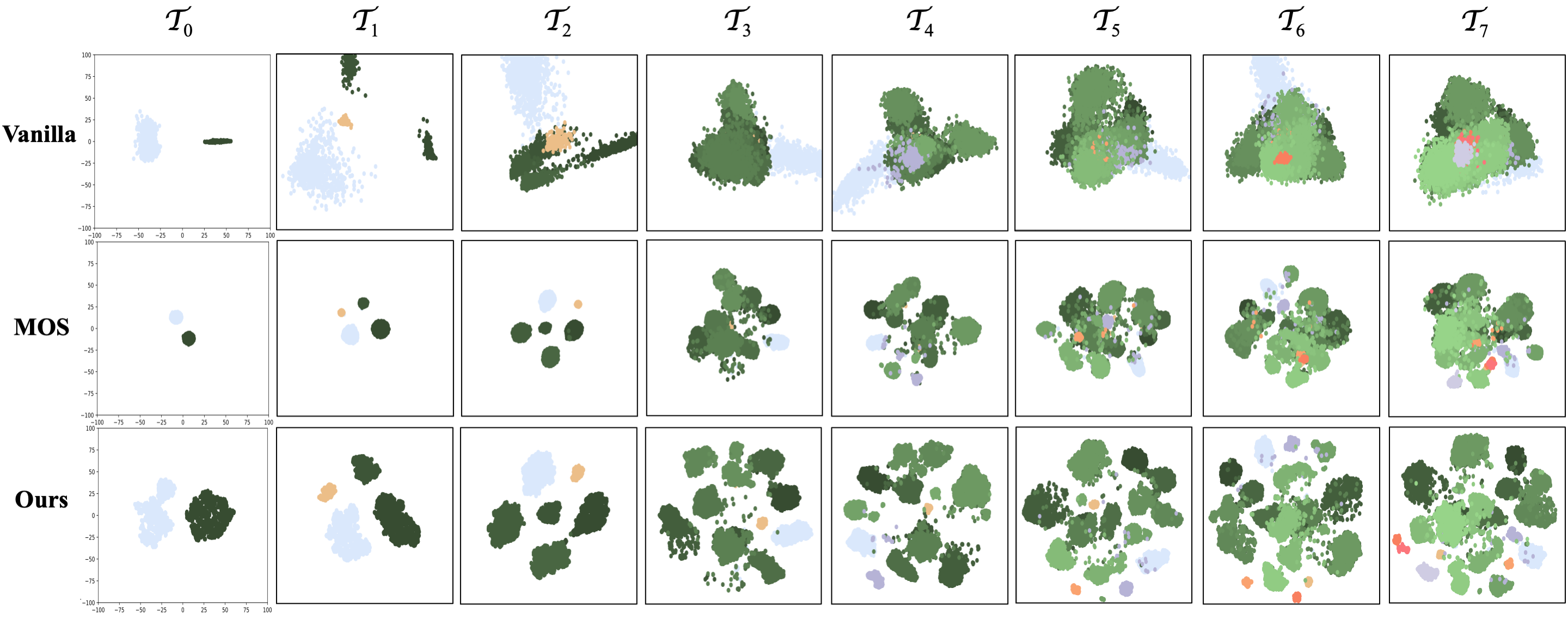}}
    \vspace{-0.1in}
    \caption{\textbf{Visualization of $z$ Across Different Tasks for CLIP, MOS, and IncreFA.} We employ t-SNE as the dimensionality reduction method, randomly selecting one class per Task. Green represents the Diffusion families, brown represents the GAN families, blue represents Real, and purple represents the AR families. We use varying shades of the corresponding colors to denote different generative models.}
    \vspace{-0.2in}
    \label{fig:tsne-vis}
\end{figure}

\paragraph{Logits Visualizations.} We visualize the outputs of three seen categories and one unseen category for comparison, as presented in \cref{fig:logits-vis}. The left in \cref{fig:logits-vis} illustrates the visualization of the softmax-applied output logits for inputs from three seen generative models, while the right corresponds to the unseen category. \(\tau\) denotes the threshold we set at 0.65. Under the constraint of \(\mathcal{L}_u\), the outputs for seen and unseen categories exhibit clear distinctions: the three seen categories display pronounced peaks exceeding the threshold, whereas the unseen category's logits are more uniformly distributed, yielding softmax probabilities all substantially below \(\tau\). This enables our method to achieve a remarkable unseen detection rate.
\vspace{-0.1in}
\paragraph{T-SNE Visualization.}  
We visualise the latent representations \(z\) of the vanilla baseline \cite{clip2021}, MOS \cite{mos2025}, and IncreFA across incremental tasks to analyse how their feature spaces evolve (\cref{fig:tsne-vis}).  
The vanilla exhibit a significant latent mixing at \(\mathcal{T}_2\). Although it can maintain a certain degree of distinction between real and fake images, as the incremental process progresses and the latent space accumulates more features, the baseline latents become almost completely mixed. The feature space of MOS exhibits clear distinguishability in the initial stages due to the presence of its adapter; however, upon introducing more latent representations, particularly with the addition of more similar families, its feature space also becomes mixed, which significantly degrades MOS's attribution performance in the later incremental stages. 
 In contrast, owing to the hierarchical constraints, IncreFA's latent space remains stable throughout, preserving the hierarchical relationships both within and between families. This constraint enables IncreFA to maintain robust attribution performance even when confronted with additional similar models.

\section{Limitations}
\label{sec:limitations}
IncreFA, though effective, has several constraints.  
The predefined family taxonomy may not generalise to ambiguous or hybrid models.  
Linear latent mixing approximates unseen distributions but cannot capture structural novelty.  
Open-set calibration still requires an empirical threshold that may drift under heavy domain shifts.  
Current experiments focus on single-image attribution; extending to multi-frame or multimodal attribution remains future work.

\section{Conclusion}
\label{sec:conclusion}
We proposed IncreFA, which reformulates image attribution as a structured incremental process.  
By encoding the hierarchical lineage of generative models and replaying compact latents for continual calibration, it achieves stable adaptation and strong open-set awareness.  
Experiments on IABench show consistent gains over prior baselines and 98.9\% unseen detection accuracy.  
The central insight is conceptual: attribution must evolve structurally with the models it traces rather than react to them.  
Future work will focus on adaptive taxonomy discovery and multimodal forensics.

\section{Acknowledgments}
This work was supported by the National Natural Science Foundation of China (Grant 62406171, 62225601, U23B2052,62306031), in part by the Beijing Natural Science Foundation Project No. L242025, in part by the  Guizhou Province Science and Technology Plan Project (No. QianKeHeZhongDa [2025]031) ,  and in part by the Fundamental Research Funds for the Beijing University of Posts and Telecommunications under Grant  2025AI4S15.

{
    \small
    \bibliographystyle{ieeenat_fullname}
    \bibliography{main}
}

\clearpage
\setcounter{page}{1}
\maketitlesupplementary

\section{More Details about IABench}
\label{sec:sup_dataset}
\begin{figure}[!t]
    \centerline{\includegraphics[width=1.0\columnwidth]{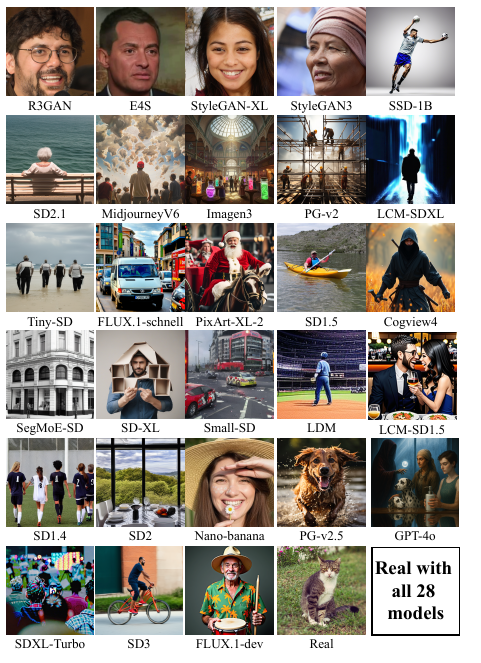}}
    \caption{\textbf{Samples visualization.}}
    \label{fig:dataset_vis}
\end{figure}

We collected generative models from 4 categories of GANs, 2 categories of autoregressive models, and 22 categories of diffusion models. The statistical information for IABench is presented in \cref{tab:dataset_tab}, including sources. The visualization of the 28 generative model categories is shown in \cref{fig:dataset_vis}. We gathered 544,333 images spanning from 2022 to 2025 and partitioned them into training and test sets. Among these, our real images are sourced from the COCO dataset. Images from CogView4 were manually generated using its official GitHub repository. 
Specifically, we employed GPT-4o to randomly generate 3,000 prompts. After removing highly similar prompts through cosine similarity comparison, we generated 7 images for each prompt using different random seeds. Following further manual screening, a total of 13,000 images were ultimately obtained.

\begin{table}[!t]
\renewcommand{\arraystretch}{0.9}
\small
\centering
\caption{\textbf{Composition of IABench}. We collected 28 categories of generative models along with real images sourced from the COCO dataset, and partitioned them into training and test sets.\note We generate all images following there official github.}
\resizebox{0.95\columnwidth}{!}{\begin{tabular}{l|crrc}
\toprule
Model                              & Release & Train  & Test & Source\\
\midrule
\textbf{Real}                      & -------      & 20000       & 5000      & \cite{cospy2025}\\
StyleGAN3 \cite{stylegan32021}                          & 02-2022      & 3600        & 900       & \cite{is2025} \\
StyleGAN-XL \cite{styleganxl2022}                        & 06-2022      & 3600        & 900       & \cite{is2025} \\
LDM \cite{ldm2022}                                & 07-2022      & 20000       & 5000      & \cite{cospy2025} \\
SD1.4 \cite{sdxl2024}                              & 08-2022      & 20000       & 5000      & \cite{cospy2025} \\
SD1.5 \cite{sdxl2024}                   & 10-2022      & 20000       & 5000      & \cite{cospy2025} \\
SD2 \cite{sdxl2024}            & 11-2022      & 20000       & 5000      & \cite{cospy2025} \\
SD2.1 \cite{sdxl2024}                      & 12-2022      & 20000       & 5000      & \cite{cospy2025} \\
Tiny-SD \cite{tinysd}                            & 06-2023      & 20000       & 5000      & \cite{cospy2025} \\
SD-XL \cite{sdxl2024}                                & 06-2023      & 20000       & 5000      & \cite{cospy2025} \\
Small-SD     \cite{sdxl2024}                      & 07-2023      & 20000       & 5000      & \cite{cospy2025} \\
SSD-1B \cite{ssd2024}                            & 07-2023      & 20000       & 5000      & \cite{cospy2025} \\
E4S \cite{e4s2023}                                & 10-2023      & 3600        & 900       & \cite{cospy2025} \\
PixArt-XL-2 \cite{pa2024}                        & 11-2023      & 20000       & 5000      & \cite{cospy2025} \\
LCM-SDXL \cite{lcm2023}                          & 11-2023      & 20000       & 5000      & \cite{cospy2025} \\
LCM-SD1.5 \cite{lcm2023}                         & 11-2023      & 20000       & 5000      & \cite{cospy2025} \\
SDXL-Turbo\cite{sdxl2024}                         & 11-2023      & 20000       & 5000      & \cite{cospy2025} \\
PG-v2 \cite{pg2024}                             & 12-2023      & 20000       & 5000      & \cite{cospy2025} \\
MidjourneyV6\cite{midjourney2023}                       & 12-2023      & 8000        & 2000      & \cite{cospy2025} \\
SegMoE-SD  \cite{sdxl2024}                        & 02-2024      & 20000       & 5000      & \cite{cospy2025} \\
GPT-4o    \cite{sharegpt2025}                         & 05-2024      & 8000        & 2000      & \cite{sharegpt2025} \\
SD3   \cite{sdxl2024}                             & 06-2024      & 20000       & 5000      & \cite{cospy2025} \\
FLUX.1-dev  \cite{flux12024}                       & 08-2024      & 20000       & 5000      & \cite{cospy2025} \\
FLUX.1-schnell   \cite{flux12024}                  & 08-2024      & 20000       & 5000      & \cite{cospy2025} \\
Imagen3      \cite{imagen32024}                     & 08-2024      & 3600        & 900       & \cite{is2025} \\
R3GAN \cite{r3gan2024}                             & 12-2024      & 3600        & 900       & \cite{is2025} \\
PG-v2.5\cite{pg2024}                         & 12-2024      & 20000       & 5000      & \cite{cospy2025}  \\
Cogview4  \cite{cogview2021}                         & 03-2025      & 10000       & 3000      & ---\note\\
Nano-banana     \cite{nano2025}                   & 08-2025      & 9844        & 3989      & \cite{nano2025} \\
\midrule
\textbf{Total}                     &              & \textbf{433844} & \textbf{110489} &  \\
\bottomrule
\end{tabular}}
\label{tab:dataset_tab}
\end{table}

\section{More Details about Experiments}
\label{sec:sup_exp}
\subsection{Incremental Baselines}
 \paragraph{iCaRL.} \cite{icarl2017} 
The model was trained for 20 epochs in the initial session with a learning rate of 0.001. Each subsequent incremental session was trained for 20 epochs with a learning rate of 0.001, decayed by a factor of 0.1 at epochs 80 and 120. The exemplar memory size was fixed at 2000 samples in total.

\paragraph{FOSTER.} \cite{foster2022} 
The model was trained for 20 epochs in the initial session with a learning rate of 0.001. Each subsequent incremental session consisted of 20 boosting epochs followed by 20 compression epochs, both using a learning rate of 0.001. The exemplar memory size was fixed at 2000 samples in total.

\paragraph{DualPrompt.} \cite{dualprompt2022} 
DualPrompt employs a pre-trained ViT-Base/16 backbone with blocks, patch embedding, cls token, norm, and positional embedding frozen. The model is trained for 5 epochs per incremental session using a constant learning rate of 0.001 and no exemplar memory. It introduces two types of learnable prompts optimized via prefix tuning: general prompts of length 5 placed at layers 0 and 1, and task-specific prompts placed at layers 2, 3, and 4. A shared batch-wise prompt pool of size 10 is maintained, where each prompt has length 5 and top-k=1 prompt is selected for each input using the CLS token as the embedding key. Prompt masking is applied, a pull constraint with coefficient 0.1 encourages proximity between selected prompts and their pool counterparts, and both the prompt pool and keys are initialized uniformly.

\paragraph{L2P.} \cite{l2p2022} 
L2P employs a pre-trained ViT-Large/16 backbone with blocks, patch embedding, cls token, norm, and positional embedding frozen. The model is trained for 5 epochs per incremental session using a constant learning rate of 0.001875 and no exemplar memory. It maintains a batch-wise prompt pool of size 10 where each prompt has length 2. For each input, the top-1 most relevant prompt is dynamically selected using the CLS token as the embedding key. A pull constraint with coefficient 0.1 is applied to regularize the selected prompts toward their corresponding pool entries. Both the prompt pool and prompt keys are initialized uniformly.

\paragraph{SimpleCIL.} \cite{simplecil2025} 
SimpleCIL employs a pre-trained ViT-Base/16 backbone using shallow visual prompt tuning with 3 learnable prompt tokens prepended to the input sequence. No exemplar memory is used. The model is trained with Adam optimizer, learning rate 0.01, weight decay 0.05, and batch size 256. The initial session with 2 classes and all subsequent incremental sessions with 4 new classes each use the same hyper-parameters and are trained until convergence with no fixed epoch count specified.
\paragraph{APER.} \cite{simplecil2025} 
APER employs a pre-trained ViT-Base/16 backbone augmented with Adapter modules. Only the adapter parameters are trained while the rest of the backbone remains frozen. The model is trained for 20 epochs per incremental session using the Adam optimizer with a learning rate of 0.001, weight decay of 0.0005, and batch size 48. No exemplar memory is used. Each adapter has a bottleneck dimension of 64.

\paragraph{DGR.} \cite{dgr2024} 
DGR combines distillation and generative replay using a ViT-Base/16 backbone. The model is trained for 20 epochs in the initial session and 20 epochs in each subsequent incremental session, both using a learning rate of 0.001 and weight decay of 0.0005. A fixed exemplar memory of 2000 samples is maintained together with a generator that produces pseudo-samples of past classes for replay training. The distillation loss weight $\lambda$ is set to 1 and the replay consistency weight $\gamma_r$ is set to 1. Temperature $\mathcal{T}$ for distillation is 2. No exemplar memory per class constraint is enforced during incremental learning.

\paragraph{TUNA.} \cite{tuna2025}
TuNA employs a pre-trained ViT-Base/16 backbone and introduces a low-rank residual adapter (r=16) into every linear layer of the transformer blocks. Only the LoRA parameters are updated during incremental learning while the original pretrained weights remain frozen. The model is trained for 15 epochs per incremental session using the Adam optimizer with cosine annealing learning rate starting at 0.01, weight decay of 0.0005, and batch size 16. A covariance-based contrastive adapter regularization with coefficient 0.001 is applied. No exemplar memory is used.

\paragraph{MOS.} \cite{mos2025} 
MOS employs a pre-trained ViT-Base/16 backbone with lightweight residual adapter modules inserted after both the attention and FFN layers of each transformer block. Only the adapter parameters are updated while the original backbone remains frozen. The model is trained for 20 epochs per incremental session using the Adam optimizer with cosine annealing learning rate starting at 0.03, weight decay of 0.0005, and batch size 48. An ensemble of the current and momentum-updated adapters with momentum 0.1 is used for inference. A covariance-based contrastive adapter regularization with coefficient 0.1 is applied. Each adapter uses a bottleneck dimension of 16. No exemplar memory is used.

\subsection{Attribution Baselines}
\paragraph{DNA-Net.} \cite{dna2022} 
DNA-Net is trained from scratch for a maximum of 30 epochs with early stopping using a batch size of 32. The optimizer uses an initial learning rate of 0.001 with exponential decay of factor 0.9 applied, applied every 2500 iterations. The contrastive loss employs a temperature of 0.07. The best model is selected based on validation accuracy.

\paragraph{RepMix.} \cite{repmix2022}
RepMix is trained for 30 epochs using the Adam optimizer with an initial learning rate of 0.0001, weight decay of 0.0005, and multi-step learning rate scheduling with gamma 0.85. The batch size is 32. Training employs a specialized mixing strategy that combines 2 images per batch (mixup samples=2) with beta 0.4 and applies mixing at layer level 5. Images are perturbed using ImageNet-C corruptions with up to 15 sequential transformations applied with probability 1.0. 

\paragraph{DE-FAKE.} \cite{defake2023} 
For DE-FAKE, we follow its original pipeline: images are captioned using BLIP, and both the generated captions and the corresponding images are encoded with the CLIP text encoder and visual encoder (ViT-B/16), respectively. The resulting 512-dimensional image and text embeddings are concatenated into a 1024-dimensional joint representation. A lightweight MLP classifier consisting of three fully-connected layers (1024 $\rightarrow{}$ 512 $\rightarrow{}$ 256 $\rightarrow{}$ 29) with ReLU activation and 0.5 dropout after the first layer is trained on top of the frozen CLIP embeddings. The entire model is trained for 50 epochs using the Adam optimizer with a learning rate of 0.0003 and cross-entropy loss. Training and testing are performed with batch size 32, and images are resized to 256×256 and normalized with mean and std of 0.5. 

\paragraph{POSE.} \cite{pose2023} 
POSE is trained from scratch for a maximum of 30 epochs using batch size 16. The classifier employs an initial learning rate of 0.0001 with exponential decay (gamma=0.9) applied every 500 iterations and uses softmax loss with temperature 0.1. An auxiliary augmentation network is jointly trained with learning rate 0.01 to generate class-specific perturbations in DCT space. Training incorporates a distance-preserving loss weighted by 0.0001, a closeness-to-known-class loss weighted by 0.01 with similarity threshold 0.95, and an MSE lower bound of 0. 

\paragraph{Siamese.} \cite{siamese2024} 
The Siamese employs an EfficientNet backbone with 512-dimensional embeddings and is trained from scratch for 50 epochs on 380×380 input images. Training follows a few-shot episodic paradigm with 3 classes per batch and 4 images per class using the Adam optimizer at a learning rate of 0.0001. 
\subsection{Additional Ablations}
\begin{table}[!t]
\begin{minipage}[!t]{0.49\columnwidth}
\centering
\caption{\textbf{Ablations on $\tau$.} We report the accuracy within seen categories and the detection rate for unseen instances following the final incremental step.}
\resizebox{0.95\linewidth}{!}{\begin{tabular}{ccc}
\toprule
$\tau$ & Seen & Unseen \\ 
\midrule
0.5                 & 78.11     & 65.01       \\
0.55                & 78.10     & 89.66       \\
0.60                & 77.34     & 95.78       \\
0.65                & {78.80}     & 98.93       \\
0.70                & 72.00     & 98.98       \\
0.75                & 69.59     & 98.53       \\
0.80                & 55.98     & 99.02       \\
0.85                & 54.23     & 98.97       \\
0.90                & 41.80     & 99.52       \\
0.95                & 39.55     & {99.79}       \\ \bottomrule
\end{tabular}}
\label{tab:tau}
\end{minipage}
\hfill
\begin{minipage}[!t]{0.49\columnwidth}
\centering
\caption{\textbf{Ablation Study on the Number of Samples per Model in the Latent Memory Bank.} We report the average accuracy following the completion of the final incremental task in the table.}
\resizebox{0.95\linewidth}{!}{\begin{tabular}{cc}
\toprule
$N_B$ per task & Final Acc. \\
\midrule
5                    & 56.04      \\
10                   & 64.68      \\
20                   & 62.32      \\
50                   & 66.30      \\
100                  & 76.99      \\
150                  & 77.80      \\
200                  & 77.82      \\
250                  & 79.23      \\
500                  & {82.00}      \\
\bottomrule
\end{tabular}}
\label{tab:bank_size}
\end{minipage}
\end{table}

\paragraph{Ablation on $\tau$.} We conducted ablation experiments on \(\tau\), with results presented in \cref{tab:tau}. When \(\tau\) is low, it implies stricter detection for unseen instances, whereas detection for seen instances becomes more lenient. Consequently, at lower levels of \(\tau\), the accuracy for unseen detection decreases substantially, while the accuracy for seen detection exhibits only minor changes. Conversely, when \(\tau\) is high, it implies stricter detection for seen instances, with any output below this threshold being classified as unseen. Consequently, as \(\tau\) increases, the accuracy for seen detection experiences a significant decline. Considering the trade-off between provenance accuracy and unseen detection performance, we set \(\tau = 0.65\) as our hyperparameter.

\paragraph{Ablation on Latent Memory Bank.}We conducted ablation experiments on different sizes of the Latent Memory Bank (denoted as \( N_B \)), with results presented in \cref{tab:bank_size}. As \( N_B \) increases, IncreFA's final average accuracy gradually improves; however, when \( N_B \) exceeds a certain threshold, the accuracy gains from further increasing \( N_B \) diminish. Clearly, a larger \(N_B\) yields higher incremental accuracy. However, in practical applications, storage costs and other constraints must be considered. Taking into account the diminishing marginal returns of increasing \(N_B\) and the actual memory footprint, we adopt \(N_B = 150\) as the final memory bank size.

\subsection{Robustness \& Threshold}
We conducted robustness experiments comparing incremental attribution and unseen Acc. across different $\tau$ reported in the table below. Our method retains reasonable performance across various degradations. 
Under degradations, $\tau$ has a stronger influence on unseen family Acc. than on incre. attribution, likely because degradation substantially increases the difficulty of attribution.
\begin{table}[!h]
\label{tab:robust}
\centering
\caption{\textbf{Experiments under different degradations.}}
\resizebox{0.98\linewidth}{!}{
\begin{tabular}{ccccccc}
\toprule
$\tau$  & JPEG-QF = 95 & QF = 90 & QF = 80 & Resizing = 128 & 256 & 384 \\
\midrule
0.60 &   74.49 / 92.17 & 67.04 / 77.45     & 68.32 / 76.52  &  71.38 / 84.83 & \textbf{72.99} / 89.71   & 75.49 / 94.92   \\
0.65 & 74.31 / 95.00          & \textbf{69.31} / 83.99     & \textbf{68.35} / 85.67     & 71.00 /  89.97    & 72.80 / 92.49  & \textbf{77.10} / 95.02   \\
0.70 &  \textbf{75.01} / \textbf{95.14}     & 68.99 / \textbf{85.43}     & 65.32 / \textbf{85.77}    & \textbf{73.85} / \textbf{93.90}          & 72.59 / \textbf{94.08}   & 74.42 / \textbf{96.00} \\
\bottomrule
\end{tabular}}
\end{table}

\subsection{Unseen \& Wrong Families}
\textbf{(a)} We conducted two experiments: one in which we masked out one known family during training and treated it as unseen during inference; the other in which we preserved the existing protocol throughout training and introduced newly emerged mixed-architecture (GLM-Image \cite{}) as unseen cases. Due to the excessive number of Diffusion models, we randomly masked out 50\% of them. We report the Avg. Acc. (Un. Acc.) across the two tasks for the unseen families scenario in the table below. Unseen families at the family level disrupt inter-model information sharing, which reasonably accounts for the observed performance drop in our method.

\begin{table}[!h]
\centering
\caption{\textbf{Experiments under masked families.}}
\resizebox{0.98\linewidth}{!}{
\begin{tabular}{ccccc}
\toprule
Unseen type & Diffusions & GANs & ARs & GLM-Image \\ 
\midrule
Increment   & 69.57 (87.29) & 75.41 (92.56) & 75.99 (95.84) & 77.94 (92.95) \\
Attribution & 90.34 (92.00) & 91.70 (94.55) &  92.08 (92.49)   &     94.67 (94.30)        \\ 
\bottomrule
\end{tabular}}
\label{tab:unseen_families}
\end{table}
\noindent\textbf{(b)} We evaluated the performance impact by intentionally assigning incorrect family labels to randomly selected sets of four generative model families in IABench. The results are reported in the table below. Similarly, incorrect families inevitably lead to performance degradation.
\begin{table}[!h]
\centering
\caption{\textbf{Experiments under wrong families.}}
\resizebox{0.98\linewidth}{!}{\begin{tabular}{ccccccc}
\toprule
Direction & D$\to$G & G$\to$D & AR$\to$G & AR$\to$D & G$\to$AR & D$\to$AR \\
\midrule
Increment   & 71.47 & 71.29 &72.53 & 71.28 & 72.56 & 73.11 \\
Attribution & 82.91 & 83.87 & 92.76 & 88.39 &   93.08     & 87.50 \\
\bottomrule
\end{tabular}}
\label{tab:wrong_families}
\end{table}

\subsection{Broader datasets}
We conduct experiments on broader datasets.  The corresponding results are reported as Acc. (Un. Acc.) in the table below.  
Due to the notably low resolution of certain samples in the RED116/140, our method experiences a clear performance degradation.
Comparisons with prior works will be supplemented in the final version.
\begin{table}[!h]
\centering
\caption{\textbf{Experiments within broader datasets.}}
\resizebox{0.98\linewidth}{!}{\begin{tabular}{ccccc}
\toprule
Datasets & RED116   & RED140   & Co.Fo. (Small)  & Scaling (GenImage + Co.Fo) \\ 
\midrule
Increment   & 65.16 (87.92) & 64.56 (84.00) & 51.05 (73.00) & 51.12 (75.39) \\
Attribution & 79.30 (96.01) & 78.91 (97.52) & 62.73 (85.19) & 64.62 (83.18) \\ 
\bottomrule
\end{tabular}}
\label{tab:more_datasets}
\end{table}

\subsection{Large Scale \& Compute Resources}
\textbf{(a)} We extended the incremental learning process on IABench by successively incorporating more generative models. The corresponding results are presented in the table below. Dataset details can be found in \textbf{Q3}. As the number of models grows, source attribution becomes increasingly difficult.
\textbf{(b)} We also compute the memory cost by fixing $N = 200$ with the results reported in the table below.
\begin{table}[!h]
\centering
\caption{\textbf{Memory cost with more models.}}
\resizebox{0.98\linewidth}{!}{\begin{tabular}{cccccc}
\toprule
Trained models & IABench & + 32 models  & + 64    & +116    &  +140     \\ 
\midrule
Incre. Acc.   & 78.80 & 72.53 & 70.00 & 64.91 & 64.19 \\
Un. Acc. & 98.93 & 98.14 & 92.20 & 89.19 & 81.20 \\ 
Memory Cost (GB) & 12.74 & 26.81 & 40.87 & 63.72 & 74.27  \\ 
\bottomrule
\end{tabular}}
\label{tab:more_incre}
\end{table}

\end{document}